\documentclass[14pt]{article}
\usepackage{graphicx}    
\usepackage{float}
\usepackage{url} 
\usepackage{multirow} 
\usepackage{amsmath,amsthm}
\usepackage{amsfonts}
\usepackage{amssymb}
\usepackage{algorithm}
\usepackage{algorithmic}
\usepackage{bbm}

\newcommand{\argmax}{\operatornamewithlimits{arg\,max}}

\newtheorem{theorem}{Theorem}
\newtheorem{lemma}[theorem]{Lemma}

\newenvironment{definition}[1][Definition]{\begin{trivlist}
\item[\hskip \labelsep {\bfseries #1}]}{\end{trivlist}}

\begin{document}         
\date{}

\title{\Large {\bf Greedy Multiple Instance Learning via Codebook Learning and Nearest Neighbor Voting}}
\author{Gang Chen, Jason Corso
\thanks{VPML, University at Buffalo, SUNY}
}
\maketitle

\begin{abstract}
Multiple instance learning (MIL) has attracted great attention recently in machine learning community. However, most MIL algorithms are very slow and cannot be applied to large datasets. In this paper, we propose a greedy strategy to speed up the multiple instance learning process. Our contribution is two fold. First, we propose a density ratio model, and show that maximizing a density ratio function is the low bound of the DD model under certain conditions. Secondly, we make use of a histogram ratio between positive bags and negative bags to represent the density ratio function and find codebooks separately for positive bags and negative bags by a greedy strategy. For testing, we make use of a nearest neighbor strategy to classify new bags. We test our method on both small benchmark datasets and the large TRECVID MED11 dataset. The experimental results show that our method yields comparable accuracy to the current state of the art, while being up to at least one order of magnitude faster. 
\end{abstract}

\section{Introduction}
Traditional supervised learning methods require a training dataset, consisting of input and label pairs, to construct a classifier that can predict outputs/labels for novel inputs. However, the requirement of input/label pairs in the training data is surprisingly prohibitive especially for large training data. Multiple instance learning (MIL) is a more flexible paradigm to learn a concept given positive and negative bags of instances. It assumes each bag may contain many instances, but a bag is labeled positive even if only one of the instances in it falls within the concept. And the bag is labeled negative only when all instances in it are negative. The aim of MIL is to induce a concept that will classify bags of instances based on the assumption defined above. This paradigm has been receiving much attention in the last several years, and has many useful applications in a number of fields, including drug activity prediction \cite{Dietterich97}, stock selection \cite{Maron98}, object detection \cite{Felzenszwalb10} and tracking \cite{Viola05}, text categorization \cite{Andrews02,Ray05} and image categorization \cite{Maron98,Zhou07}. 

Although MIL has received an increasing amount of attention in recent years, the problem is still fairly undeveloped and there are many interesting open questions. The MIL problem is harder than traditional supervised learning methods because the learner receives a set of bags instead of a set of instances that are labeled positive or negative. Moreover, the learner needs to deal with the false positives in positive bags. Recently research \cite{Kundakcioglu10} also shows that the halfspaces finding problem for MIL is NP-complete. And most current MIL algorithms are still very slow and cannot be applied to large data sets. Thus, an approximate algorithm to efficiently find implicit or explicit decision boundary is vital to solve the MIL problem.  

\begin{figure}[ht!]
\centering
\includegraphics[trim = 15mm 86mm 32mm 60mm,clip=true, width=18.0cm]{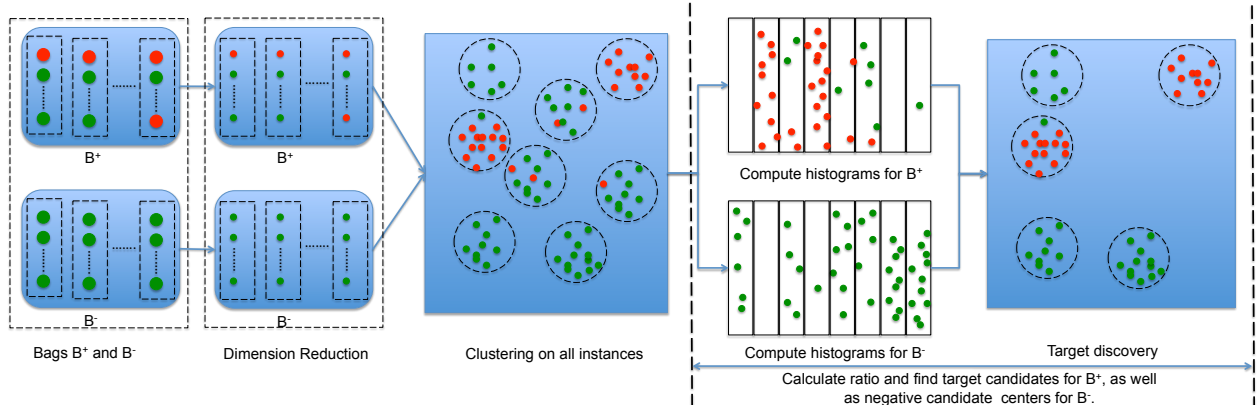}
\caption{An overview of our training process to find targets of positive bags, and negative candidate centers of negative bags. First, for all instances, we reduce the high feature space to low dimension by using PCA (optional step) for feature selection; then we cluster all instances from both positive bags and negative bags into $K$ clusters. Next, we count instances and bin them into $K$ centers for positive bags and negative bags separately. Lastly, by computing the histogram ratio between positive bins and negative bins, we rank each bin to discover target codebooks and negative candidate centers. These discriminative codebooks can be used to classify new bags through a nearest neighbor strategy.}
\label{F1}
\end{figure}

In this paper, we propose a greedy multiple instance learning method (GMIL) via codebook learning and nearest neighbor voting. Our approach is inspired by the definition of training bags, as well as the Diverse Density (DD) \cite{Maron98} and Citation kNN models \cite{Wang00}. If there's one true concept $t$ in positive bags, it must be the intersection of all positive bags and it must be far from negative bags. In other words, it has higher density on positive bags and lower density on negative bags. Thus, we present the density ratio model and derive the relationship between the DD model and our method. Instead of maximizing a likelihood function, we maximize a density ratio between positive bags and negative bags. Then, we take a greedy strategy to select codebooks for positive targets and negative candidates separately by sorting the histogram ratio between positive bags and negative bags. Our algorithm is very fast, which is linear in the number of total instances from all training bags. See Fig. (\ref{F1}) for the training process. As for classification, we take a nearest neighbor strategy. We test our method on benchmark dataset and the TRECVID MED11 dataset, and our results demonstrate that our method is comparable to the state of the art. 

The rest of the paper is structured as follow: In Section 2, we surveyed the related work. We describe the the greedy multiple instance learning algorithm in Section 3. Details of the experimental setup and the results of the experiment are given in Section 4, and we conclude in Section 5.

\section{Related work}
One of the earliest algorithms for learning from multiple instances was developed by Dietterich et al. \cite{Dietterich97} for drug activity prediction.  Their algorithm, the axis-parallel rectangle (APR) method, expands or shrinks a hyper-rectangle in the instance feature space with the goal of finding the smallest box that covers at least one instance from each positive bag and no instances from any negative bag. 
Following this seminal work, there has been a significant amount of research devoted to MIL problems using different learning models, such as DD \cite{Maron98}, EM-DD \cite{Zhang01}, and extended Citation kNN \cite{Wang00}. 

Due to the success of the SVM algorithm \cite{Andrews02}, and the various positive theoretical results behind it, maximum margin methods have become extremely popular in machine learning. Moreover, to improve classification accuracy, many variations of SVM have been proposed by changing constraints, the objective functions, space projection, kernels, etc. Andrews et al. \cite{Andrews02} combined MIL with SVM first to cope with MIL problems. Later methods focus more on applying discriminative models to solve MIL problems, such as SVM \cite{Bunescu07,Chen06, Cheung06,Gehler07,Tao04,Zhou09}, neural networks \cite{Keeler90,Ramon00}, boosting \cite{Andrews04,Viola05}, regression for feature selection \cite{Ray05,Raykar08}, decision trees \cite{Yann01}, mixtures of Gaussian \cite{Wang08}, Gaussian processes \cite{Kim10}, conditional random fields \cite{Deselaers10} and manifold learning \cite{Babenko11}. To our knowledge, most work on MIL has paid little attention to efficiency or testing MIL approaches on large datasets. There is also no work on learning codebooks for both positive bags and negative bags.
As mentioned in \cite{Maron98}, the assumption that all bags intersect at a single point is not necessary, DD assumes more complicated concepts. However, it was not generalized to learn negative targets to deal with false positives. Recently, discriminative dictionary learning \cite{Mairal08} greatly improves accuracy for visual object recognition. Thus, learning a good dictionary is vital for a discriminative approach. We argue that learning discriminative targets, including both positive and negative centers, can yield a better representation for training data. But how to choose discriminative clusters for MIL problems? In this paper, we propose a greedy multiple instance learning via codebook learning and nearest neighbor voting. We introduce the density ratio function which is low bound of the DD model. We also generalize DD model to learn both positive and negative centers, which will help reduce the false positives in classification. Moreover, our method is very fast for training, which is linear in the number of all instances, at least an order of magnitude faster than comparable methods, and can be applied to large dataset.

\section{Greedy multiple instance learning}
In this section, we present a greedy strategy for multiple instance learning. We maximize the density (histogram) ratio between positive bags and negative bags to find codebooks for both targets and negative candidate centers, which can be used to classify novel bags. The notation used in this paper for bags and instances is the one introduced by Maron and Lozano-Perez \cite{Maron98}. 

\subsection{Density ratio function}
Let $D$ be the labeled data which consists of a set of positive bags and negative bags. We denote all positive bags as $B^{+}$ and all negative bags as $B^{-}$. We denote $B_{i}^{+}$ as the $i^{th}$ positive bag, the $j^{th}$ point (instance) in that bag as $B_{ij}^{+}$, and the value of the $k^{th}$ feature of that point as $B_{ijk}^{+}$. Likewise, $B_{ij}^{-}$ represents a negative point. For training data $D = \{B_{1}^{+},...,B_{n}^{+}; B_{1}^{-},...,B_{m}^{-}\}$, assuming for now that the true concept is a single point $t$ (using Bayes' rule and assuming i.i.d for observations and a uniform prior over concept location). Thus, to find the true concept is equivalent to finding a point, which has high density on positive bags, and low density on negative bags. As a result, we maximize the following density ratio:
\begin{eqnarray}\label{eq:Eq1}
\begin{aligned}
\quad  \underset{x}{\operatorname{arg\,max}}  \frac {\sum_{i=1}^n \sum_{j} Pr(x=t \mid B_{ij}^{+} )} {\sum_{i=1}^m \sum_{j} Pr(x=t \mid B_{ij}^{-})} \\
\end{aligned}
\end{eqnarray}
Then, we can employ kernel density estimation strategy \cite{duda2001} to calculate the probability at the point $x$, and rewrite Eq. (\ref{eq:Eq1}) as the following
\begin{eqnarray}\label{eq:Eq2}
\begin{aligned}
\centering
&\argmax_{x} \frac{\sum_{i=1}^n\sum_{j} \varphi(\frac{x-t}{h_{n}}|B_{ij}^{+})}{\sum_{i=1}^m\sum_{j} \varphi(\frac{x-t}{h_{n}}|B_{ij}^{-})} &\\
\end{aligned}
\end{eqnarray}
where $h_{n}$ is the edge length in a $d$-dimension volume, $\varphi$ is a Parzen windowing function defined on $d$-dimension vector $u$:
\[ \varphi (u) =\begin{cases}
1 & |u_{j}|<\frac{1}{2}, j=1,...,d\\
0 & otherwise
\end{cases} \]
Note that we ignore the constants in Eq. (\ref{eq:Eq2}). We also argue that we can find the true concept by maximizing Eq. (\ref{eq:Eq2}). Intuitively, if one of the instances in a positive bag is close to $t$, then $\varphi(\frac{x-t}{h_{n}}|B_{ij}^{+}) \approx 1$, and $\varphi(\frac{x-t}{h_{n}}|B_{ij}^{-}) \approx 0$. Further, if every positive bag has an instance $x$ near to $t$ and no negative bags are close to $t$, then $x$ will have high density in Eq. (\ref{eq:Eq2}). Let us consider the case where $\forall x\in B_{i}^{+}$ is not close to the true concept $t$; then the numerator will be approximately zero, but the denominator will be larger than zero because it must fall into negative bags. If $x$ is close to the target, it means the denominator will be approximately zero, while the numerator will approximate the total number of positive bags because each of them contains at least one positive instances. Hence, 
 Eq. (\ref{eq:Eq2}) can help find the target hypothesis $t$.
 
However, it is always difficult to define the length of an edge $h_{n}$ in a $d$-dimension hypercube space. In addition, maximizing Eq. (\ref{eq:Eq2}) is a tough problem because it is not a continuous and differentiable function. We can use a brute force searching strategy to find the target, but it is time-consuming. Because all positive bags contains at least one instance $x$ close to $t$, and no negative bags are close to $t$, thus we can make use of clustering methods to allocate all instances into $K$ bins. Then, we can approximately calculate Eq. (\ref{eq:Eq2}) by maximizing the ratio between positive and negative histograms.

We make use of K-means to cluster all instances into $K$ clusters, denoted as $C_{1}, C_{2}, ...,C_{K}$. For positive bags, we calculate the following and count the instances (frequency) falling into $K$ bins
\begin{equation}\label{eq:Eq3}
\centering
bin(k)^{+} = {\#}(B_{ij}^{+} \in C_{k}), k=1,...,K
\end{equation} 
By normalizing over all $K$ bins, we can get the histogram for $B^{+}$ as: $h(k)^{+} =\frac{bin(k)^{+}}{\sum_{k=1}^{K}bin(k)^{+}}$.

Likewise, we can count instances falling into $K$ bins for negative bags
\begin{equation}\label{eq:Eq4}
\centering
bin(k)^{-} =\#(B_{ij}^{-} \in C_{k}), k=1,...,K
\end{equation} 
Similarly, we can calculate the histogram for $B^{-}$ as: $h(k)^{-} =\frac{bin(k)^{-}}{\sum_{k=1}^{K}bin(k)^{-}}$, $k=1,...,K$.

Thus, we can make use of the following formula to approximate Eq. (\ref{eq:Eq2})
\begin{equation}\label{eq:Eq5}
\centering
\argmax_{x} \frac{\sum_{i=1}^n\sum_{j} \varphi(\frac{x-t}{h_{n}}|B_{ij}^{+})}{\sum_{i=1}^m\sum_{j} \varphi(\frac{x-t}{h_{n}}|B_{ij}^{-})}
\propto \argmax_{k} \frac{h(k)^{+}}{h(k)^{-}} 
\end{equation}
where $k = 1,...,K$.
\subsection{Target codebook discovery}
It is possible that the denominator in Eq. (\ref{eq:Eq5}) might be equal to 0. We introduce a small positive constant $\epsilon$ to avoid such a situation. In addition, we want larger $bin(k)^{+}$ in positive bags to have higher priority chosen as target centers; we introduce the Sigmoid function as weights and reformulate Eq. (\ref{eq:Eq5}) as follows:
\begin{equation}\label{eq:Eq6}
\centering
\argmax_{k} \frac{h(k)^{+}}{h(k)^{-}+\epsilon}\times \sigma(\frac{bin(k)^{+}-n/K}{n/K}), k=1,...,K
\end{equation}
where $\sigma(x) = \frac{1}{1+e^{-x}}$, and $n$ is the total number of positive bags $B^{+}$. Remember that $n/K$ is just the average number of positive instances in each bin if there is only one positive instance in each positive bag. Roughly speaking, if $x$ is the intersection of $n$ bags, it should aggregate into one bin, and that bin's frequency should be larger than $n/K$. Thus, we want to increase the weights of bins which have higher numbers of instances from positive bags. It is straightforward to choose $k$ that maximize Eq. (\ref{eq:Eq6}) for one target point. Furthermore, rather than having just one target point $t$, it is also straightforward to select  the second center (or bin) with the next largest histogram ratio. To find more target candidates, one can sort the ratio of each bin from largest to smallest in Eq. (\ref{eq:Eq6}), and then greedily select the largest for example \textbf{p} centers as the target codebooks, denoted as $C^{+}$ = $\{C_{i}^{+} \mid i=1,...,\textbf{p}\}$, where $\textbf{p}\geq 1$. Note that such greedy strategy to search codebooks makes sure our targets appear with higher probability in positive bags and lower probability in negative bags.  

As for negative candidate centers, we take a different strategy. Note that since all instances from negative bags are negative, it means that the most counted cluster centers in Eq. (\ref{eq:Eq4}) can be used as the codebooks for all negative instances. In this paper, we sort the negative histogram $h(k)^{-}$ descendingly, with $k=1,...,K$, and choose the first \textbf{q} centers as the negative representatives, denoted as $C^{-}$ = $\{C_{i}^{-} \mid i=1,...,\textbf{q}\}$, where $\textbf{q}\geq 0$. In a sense, our approach generalizes the DD model to learning both positive and negative targets.

\subsection{Nearest neighbor for classification}
Assuming we have learned data centers: $C^{+}$ with \textbf{p} positive centers and $C^{-}$ with \textbf{q} negative centers, we employ nearest neighbor to classify new bags. For each new bag, we calculate its distance to both \textbf{p} positive centers and \textbf{q} negative centers, and find the minimum \textit{Hausdorff distance} \cite{RockafellarWets98} as the distance between the bag and the learned codebooks. 

Let us consider two situations below. (1) Assume every positive bag shares only one target, namely $\textbf{p} = 1$ and $\textbf{q} = 0$. In this situation, we set a threshold $\tau$ to measure ``closeness'' for a new unlabeled bag $B$. In this paper, we define $\tau$ as the mean distance from input bags to the target. Such a thresholding strategy is similar to the probability threshold in the DD model. (2) Note that all bags intersecting at a single point is not necessary. We can assume more complicated concepts, for example, $\textbf{p} \geq 1$ and $\textbf{q} \geq 1$.
For a new bag $B$ with instances \textbf{x} = \{$x_{i}, i =1,...,n$\}, we label the bag 
\begin{equation}\label{eq:Eq7}
label(B) =\begin{cases}
1 & dist(\textbf{x},C^{+})<dist(\textbf{x},C^{-})\\
0 & otherwise
\end{cases}
\end{equation}
where $dist(\textbf{x},\textbf{y})$ is the minimum \textit{Hausdorff} distance between two sets $\textbf{x}$ and $\textbf{y}$. $C^{+}$ and $C^{-}$ are positive and negative centers respectively. 

We prefer kNN to SVM, although kNNs have similar ``behavior" to SVMs (both of them need to learn an explicit or implicit decision boundary from training data). Since, we have learned discriminative centers, we can directly apply kNN for classification. Additionally, kNN has good Bayesian bound; the error rate of a kNN classifier tends to the Bayes optimal theoretically as the sample size tends to infinity.

\subsection{Relationship with Diverse Density model}
For one target hypothesis problem, we derive that the density ratio model is the low bound of the Diverse Density model.

\begin{definition}
For two classes problem with balanced training data ($m=n$), if the positive training data is separable from the negative ones in Euclidean space, then we say the training data is well distributed. Furthermore, if the desired target in positive bags is separable from all negative instances (contained in both positive and negative bags), we say the multiple instance learning problem is well distributed.
\end{definition}
One of the well known example is the Gaussian distribution. In order to find the desired target, in other words the intersection of the positive bags, we hope it obeys Gaussian distribution and is separable from the negative instances.
\begin{lemma}
\label{LeftCosetsDisjoint}
If the multiple instance learning problem is well distributed, then for $\forall x \in B^{+}$ that is close to the target $t$, we have $\varphi(\frac{x-t}{h_{n}}|B_{ij}^{+}) \ge \varphi(\frac{x-t}{h_{n}}|B_{ij}^{-})$.
\end{lemma}
This is straightforward from the definition.
\begin{theorem}\label{th1}
For the $2n$ variables $v_{i}^{+}, v_{i}^{-} \in (0,1]$, $i =1,..., n$, if $v_{i}^{+}\ge v_{j}^{-}, \forall i, j \in [1, n]$, then we have $\frac{\sum_{i=1}^{n} v_{i}}{\sum_{i=n+1}^{2n} v_{i}} \le \frac{\prod_{i=1}^{n} v_{i}}{\prod_{i=n+1}^{2n} v_{i}}$. 
\begin{proof}
We only prove this for n=2, and it is easy to extend it to more general situation. Assume we have only 4 variable, $v_{1}^{+}, v_{2}^{+}$, $v_{1}^{-}$ and $v_{2}^{-}$, where $v_{i}^{+}\ge v_{j}^{-}$, $i, j \in [1, 2]$, then we have to prove $\frac{v_{1}^{+}v_{2}^{+}}{v_{1}^{-}v_{2}^{-}} \ge \frac{v_{1}+ v_{2}}{v_{3}+v_{4}}$.
\begin{align}
&\frac{v_{1}^{+}v_{2}^{+}}{v_{1}^{-}v_{2}^{-}} - \frac{v_{1}^{+}+ v_{2}^{+}}{v_{1}^{-}+v_{2}^{-}}& \\
= & \frac{v_{1}^{+}v_{2}^{+}(v_{1}^{-}+v_{2}^{-}) - v_{1}^{-}v_{2}^{-}(v_{1}^{+}+v_{2}^{+})}{v_{1}^{-}v_{2}^{-}(v_{1}^{-}+v_{2}^{-})}& \\
=&\frac{v_{2}^{+}v_{1}^{-}(v_{1}^{+}-v_{2}^{-}) + v_{1}^{+}v_{2}^{-}(v_{2}^{+}-v_{1}^{-})}{v_{1}^{-}v_{2}^{-}(v_{1}^{-}+v_{2}^{-})}& \\
\ge& 0&
\end{align}
\end{proof}
\end{theorem}
\begin{lemma}
For well distributed MIL problem, if there are balanced training number of bags, and further $\forall i$, $\sum_{j} \varphi(\frac{x-t}{h_{n}}|B_{ij}^{+}) \ge \sum_{j} \varphi(\frac{x-t}{h_{n}}|B_{ij}^{-})$, then density ratio is the low bound of Diverse density model.
\end{lemma}
Refer Appendices for the proof. Note that the condition $\forall i$, $\sum_{j} \varphi(\frac{x-t}{h_{n}}|B_{ij}^{+}) \ge \sum_{j} \varphi(\frac{x-t}{h_{n}}|B_{ij}^{-})$ is very weak. Because there's at least one positive instance in positive bags, thus $\exists x \in B_{i}^{+}$ close to the target $t$, such that $\varphi(\frac{x-t}{h_{n}}|B_{ij}^{+}) \approx 1$. While $\forall x \in B_{i}^{-}$, it is far from $t$, thus $\varphi(\frac{x-t}{h_{n}}|B_{ij}^{-}) \approx 0$. According to such analysis, we can conclude the condition is weak.
\subsection{Algorithm}
We summarize the above discussion in pseudo code. Considering that the traditional K-means depends on the initial clusters, we use more robust K-means++ \cite{Arthur07} in our experiment. 
\subsection{Complexity}
Suppose we have a total $n$ positive bags and $m$ negative bags, with $N$ total instances. The complexity of our algorithm consists of K-means clustering and computing histograms for both positive bags and negative bags. The other steps for training only operate on constant numbers, so their corresponding time can be ignored. Note that the complexity of our method is dominated by K-means, which can be finished in $O(NK)$. For large datasets, we randomly sample 10000 instances from the training data, and employ K-means++ to partition them into $K$ clusters, and then assign all other instances into the $K$ centers.
 \begin{algorithm}[t!]
\caption{} 
\label{alg1} 
\begin{algorithmic}[1]
\STATE Initialize $K$, $\epsilon$, $\textbf{p}$, $\textbf{q}$, $\tau$;
\STATE Partition $D={D_{1}, D_{2},..., D_{10}}$; // 10-fold cross validation
\FOR{$i=1;  i<=10;  i++$}
\STATE $D_{t} = D- D_{i}$; //$D_{t}$ training data, $D_{i}$ for testing
\STATE Do k-means++ and cluster $D_{t}$ into $K$ bins;
\STATE Count instances which fall into $K$ bins from $B^{+}$ using Eq. (\ref{eq:Eq3});
\STATE Normalize items in $K$ bins from $B^{+}$;// compute histogram for $B^{+}$
\STATE Count instances which fall into $K$ bins from $B^{-}$ using Eq. (\ref{eq:Eq4});
\STATE Normalize items in $K$ bins from $B^{-}$; // compute histogram for $B^{-}$
\STATE Compute histogram ratio between $B^{+}$ and $B^{-}$ according to Eq. (\ref{eq:Eq6});
\STATE Find the first \textbf{p} targets as positive codebook according to Eq. (\ref{eq:Eq6});
\STATE Find the first \textbf{q} negative candidate centers as negative codebook according to Eq. (\ref{eq:Eq4});
\STATE Compute accuracy for $D_{i}$ by nearest neighbor strategy;
\ENDFOR
\STATE Return average accuracy;
\end{algorithmic}
\end{algorithm}

\section{Experiments}
We conducted experimental evaluation on five benchmark datasets including the traditional MUSK datasets (Musk1 and Musk2) \cite{Dietterich97} and image datasets (Tiger, Elephant, and Fox)\footnote{\url{http://www.cs.columbia.edu/~andrews/mil/datasets.html}}. We also evaluate our method on a large dataset, TRECVID MED11 dataset\footnote{\url{http://www.nist.gov/itl/iad/mig/med11.cfm}}. Ten-fold cross-validation was used and the per-fold average test classification performance was calculated for evaluation. For parameter settings, in general, $K$ is related to dataset size. Larger dataset, larger $K$. $\tau$ is related to variance of each cluster after K-means clustering. $\textbf{p}$ and $\textbf{q}$ are decided by the numbers of the positive bags and negative bags. Specifically, we set $\textbf{p} \ge \textbf{q}$ for balanced training dataset, and $\textbf{p} \le \textbf{q}$ for unbalanced training data (more negative data). All parameters in the experiments are determined empirically.

\subsection{Benchmark MIL datasets}
\begin{figure}[h!]
\begin{center}
\begin{tabular}{cc}
\includegraphics[trim = 38mm 84mm 42mm 85mm,clip=true, width=5.8cm]{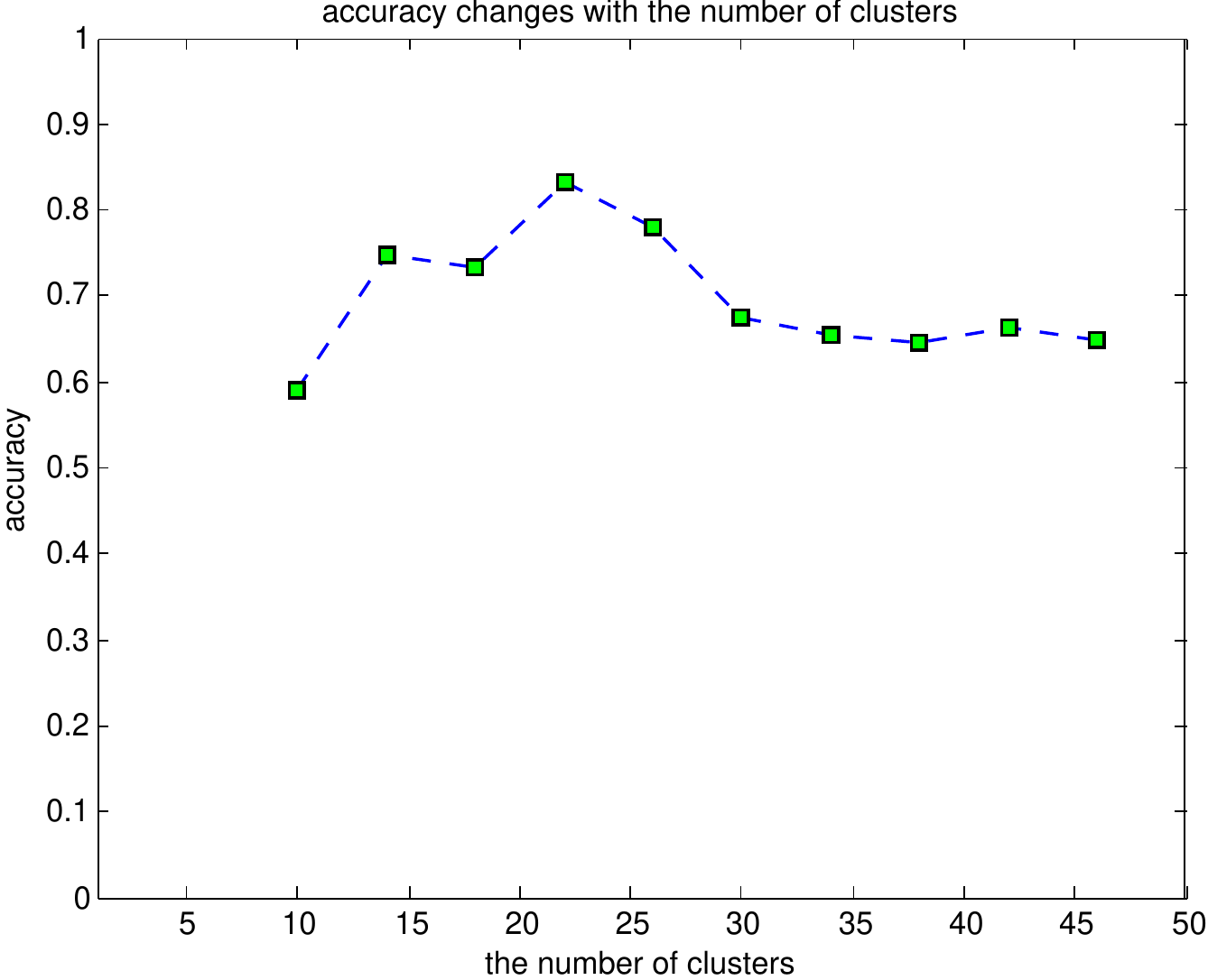}&
\includegraphics[trim = 38mm 84mm 42mm 85mm,clip=true, width=5.8cm]{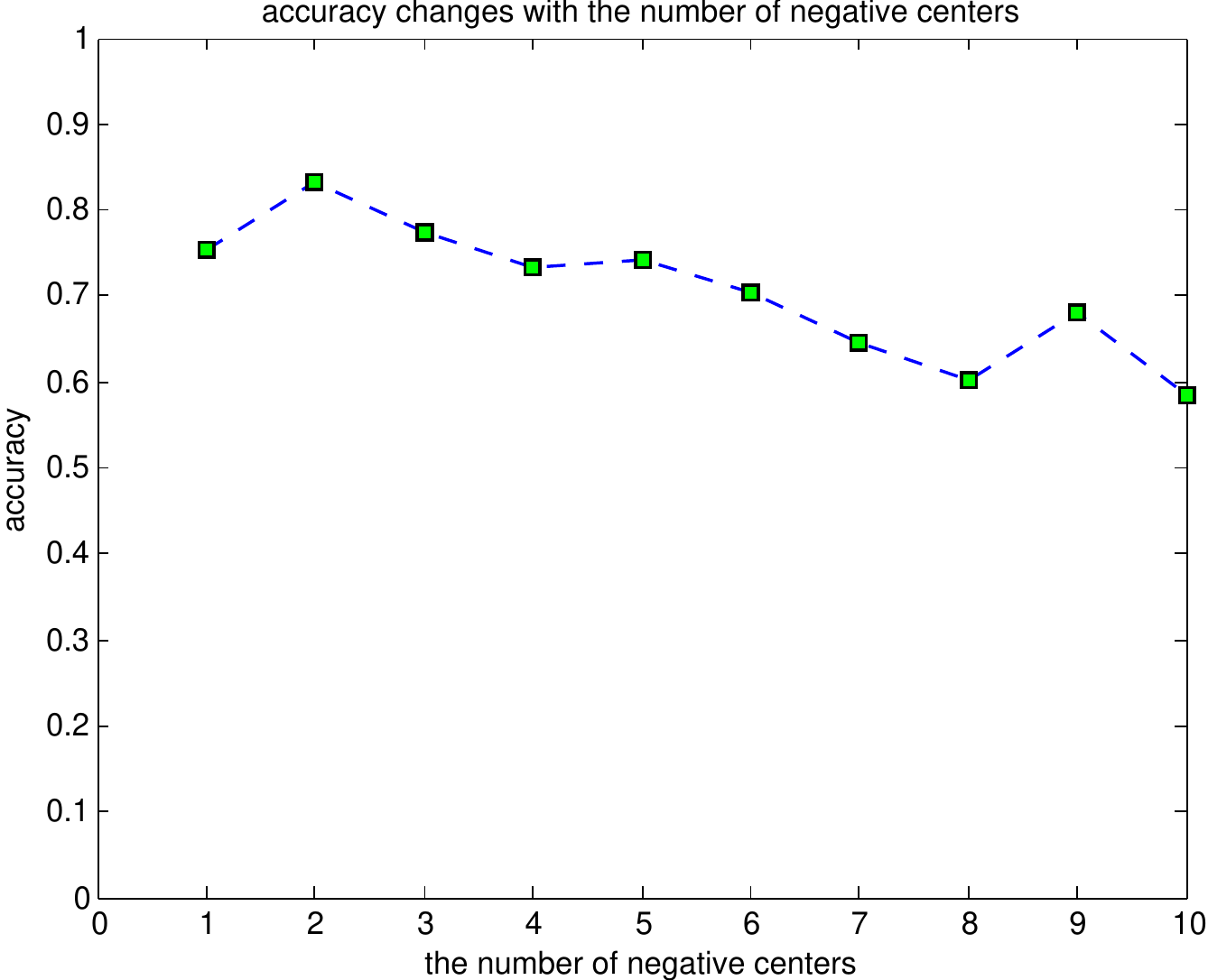}\\
(a) & (b)
\end{tabular}
\end{center}
\caption{(a) We fixed $\textbf{p} = 2$ and $\textbf{q} = 2$, and vary $K$. It describes accuracy changes with the number of clusters for Musk2 dataset. (b) We fix $K=22$, $\textbf{p} = 2$, and vary the number of negative centers $\textbf{q}$ for Musk2 dataset. It shows that learning negative codebooks is useful, but too large $\textbf{q}$ will decrease accuracy.}
\label{fig:Fig7}
\end{figure}

The MUSK datasets have widely served as the benchmark dataset for the MIL algorithms.  The feature vector for both Musk1 and Musk2 is 166-dimensional. The MUSK1 contains a total of 92 bags (47 positive and 45 negative), with approximately 6 instances per bag. The Musk2 dataset contains 102 bags (39 positive/63 negative), with 65 instances per bag on average. The COREL image dataset is 230 dimensional, containing three object categories: tiger, elephant, and fox. Each of the three categories consist of 200 bags (100 positive and 100 negative), with about 6 instances per bag. We compare our method with many others, ranging from classical MIL algorithms (the Matlab codes\footnote{\url{http://www.cs.cmu.edu/~juny/MILL/}}  of DD, EM-DD, Citation-KNN and mi-SVM are available) to recent models (miGraph\footnote{\url{http://lamda.nju.edu.cn/Data.ashx}}, CRF-MIL, GPMIL) in multiple instance learning. Tab. (\ref{Tab1}) summarizes the performance of twelve MIL algorithms in the literature.
There are five parameters $K$, $\epsilon$, $\textbf{p}$, $\textbf{q}$, $\tau$ need to be specified for our method. We set $K=10$, $\textbf{p} =2$, $\textbf{q} =0$ and $\tau = 0.082$ for Musk1 dataset. As for Musk2 dataset, we set $K=22$, $\textbf{p} =2$, $\textbf{q} =2$. In order to describe how $K$, \textbf{p} and \textbf{q} influence accuracy, we vary $K$ and $q$ respectively on Musk2 dataset. We demonstrate that learning negative codebooks is necessary, while too large $\textbf{q}$ will decrease accuracy, see Fig. (\ref{fig:Fig7}) for an intuitive understanding. As for COREL image dataset, we set $K=20$, $\textbf{p} =4$, $\textbf{q} =4$ for all three objects. Our method can yield accuracy that is comparable to most MIL methods on the benchmark dataset. For the fox dataset, our highest accuracy is 70\%, which is higher than previous methods. See Table. (\ref{Tab1}). The comparisons of different methods demonstrate again that no single MIL algorithm outperforms the others across all data sets \cite{Ray05}. 

\begin{table*}[ht!]
\begin{center}
\begin{tabular}{lcccccc}
\hline
\multicolumn{1}{l}{\multirow{2}{*}{Method}}& \multicolumn{5}{c}{Accuracy for benchmark dataset} \\
\cline{2-6}
& Musk1 & Musk2 & Elephant & Fox & Tiger\\
\cline{1-6}
APR\cite{Dietterich97} & 92.4 & 89.2 &  N/A & N/A & N/A\\
DD\cite{Maron98} & 88.9 & 82.5 & N/A & N/A & N/A\\
EM-DD\cite{Zhang01} & 84.8 & 84.9 & 78.3  & 56.1  & 72.1\\
Citation kNN\cite{Wang00} & 92.0 & 86.3 & N/A & N/A & N/A\\
mi-SVM\cite{Andrews02} & 87.4 & 86.3 & 80.0 & 57.9 & 78.9\\
MICA\cite{Fung06} &  84.4 &  $\textbf{90.5}$ & 82.5 & 62.0 & 82.0\\
MI-SVM + DA\cite{Gehler07}  & 85.7  &  83.8 & 82.0 & 63.5 & 83.0 \\
PPMM Kernel\cite{Wang08}  & $\textbf{95.6}$  & 81.2 & 82.4 & 60.3 & 80.2\\
miGraph\cite{Zhou09}& 90.0 & 90.0 & $\textbf{85.1}$ & 61.2 & 81.9 \\
CRF-MIL\cite{Deselaers10} & 88.0 & 85.3 & 85.0 & 67.5 & 83.0 \\
GP-MIL\cite{Kim10}& 89.5 & 87.2 & 83.8 & 65.7 & $\textbf{87.4}$ \\
GMIL (this paper) & 85.0 & 83.2 & 81.0 & $\textbf{70.0}$ & 82.0 \\
\hline
\end{tabular}
\caption{Accuracy comparison on five standard datasets for different methods. The results of other methods are taken from the respective papers. It demonstrates that our method can yield comparable results. Note that except Citation kNN \cite{Wang00}, the results of all other methods are based on 10 fold cross-validation.}
\label{Tab1}
\end{center}
\end{table*}


We also compare efficiency between different methods conducted in the same environment (PC + Matlab2010a). The efficiency of an algorithm was measured based on the running time, i.e., the CPU time measured in seconds until the algorithm meets the stopping criterion. Note that in order to speed up the DD and EM-DD algorithms, we set parameters, such as iterations and F-count for Matlab function \textit{fmincon} smaller than default value in the source code. We also test the MIL-Ensemble source code\footnote{\url{http://lamda.nju.edu.cn/code_MIL-Ensemble.ashx}}. Because it is equally slow as DD, we do not include its results in Table  (\ref{Tab2}). A significant advantage of our method is its speed: it is faster than all other methods by at least an order of magnitude. 
\begin{table*}
\begin{center}
\begin{tabular}{lccccccc}
\hline
\multicolumn{1}{l}{\multirow{2}{*}{Method}}& \multicolumn{6}{c}{Time (sec) for different dataset} \\
\cline{2-8}
& Musk1 & Musk2 & Elephant & Fox & Tiger & Avg. rank & Code\\
\cline{1-8}
DD \cite{Maron98} & 1725.9 & $>$3600 &  $>$3600 &  $>$3600 & $>$3600  & - & Matlab\\
EM-DD \cite{Zhang01} & 1693.3 & $>$3600 & $>$3600 & $>$3600 & $>$3600 & - & Matlab\\
Citation kNN \cite{Wang00}  & 18.1 & 2920.9 & 153.9 & 139.1 & 120.5  & 4 & Matlab\\
MI-SVM \cite{Andrews02} & 3.2 & 281.8 & 22.1 & 31.4 & 32.1 & 2 & Matlab/C++\\
miGraph \cite{Zhou09} & 21.3 & 150.1 & 102.9 & 102.4 & 101.4 & 3 & Matlab/C++\\
CRF-MIL\cite{Deselaers10} & 200.0 & N/A & N/A & N/A & N/A & - & Matlab\\
GMIL (Our method) & 1.7 & 13.7 & 11.9 & 8.1 & 9.6 & 1 & Matlab\\
\hline
\end{tabular}
\caption{Time required for 10-fold cross validation by different methods with source code available, except CRF-MIL. Note that CRF-MIL's running time for Musk1 was taken in the corresponding paper. This experiment result was based on a PC with 1 Intel Core 2 Duo CPU(2.26 GHz), 4GB RAM and MATLAB R2010a.}
\label{Tab2}
\end{center}
\end{table*}
Overall our method is ranked fifth among those in the evaluation, making it comparable to the other global techniques. However these other techniques all run one order of magnitude more slowly than our method. It is also important to note that our results are based only on the simple K-means method, whereas other methods use additional information as well as more sophisticated data costs. We used simple K-means with Euclidean distance because our focus here is on the algorithmic techniques, and demonstrating that our method produces similar quality results much more quickly. 
\subsection{Experiments on the TRECVID MED11}
\begin{figure*}[t!]
\centering
\begin{tabular}{ccccc}
\includegraphics[trim = 10mm 5mm 10mm 5mm,clip=true, width=3.1cm, height = 1.6cm]{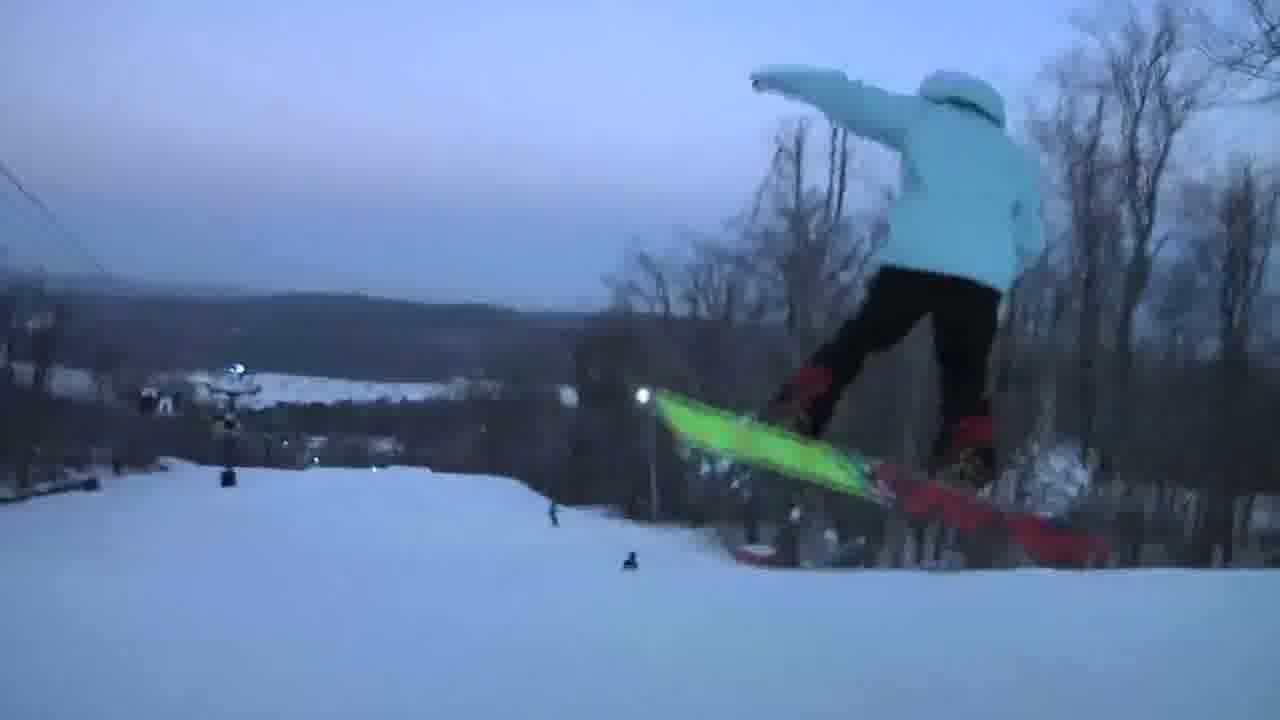}&
\includegraphics[trim = 10mm 5mm 20mm 5mm,clip=true, width=2.6cm, height=1.8cm]{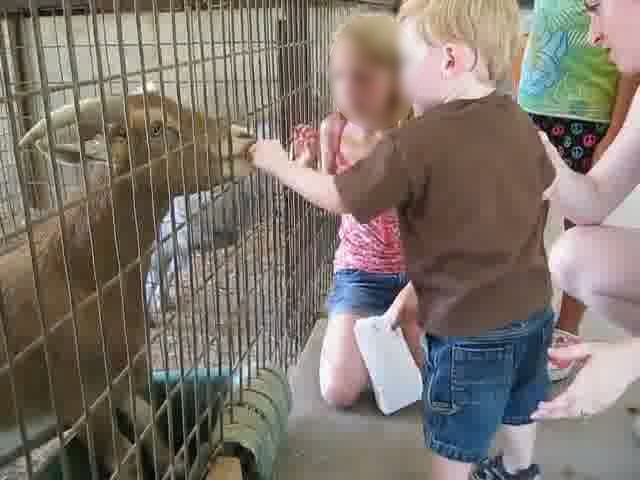}&
\includegraphics[trim = 10mm 5mm 20mm 5mm,clip=true, width=2.6cm, height = 1.8cm]{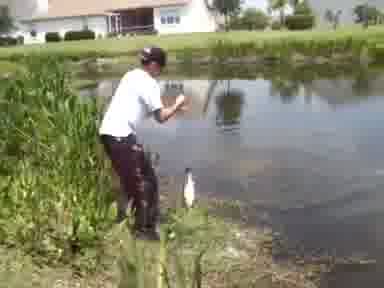}&
\includegraphics[trim = 10mm 5mm 20mm 5mm,clip=true, width=2.6cm, height = 1.8cm]{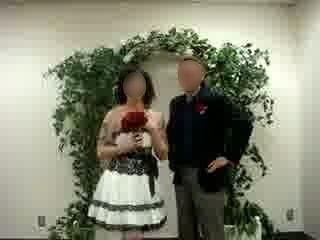}&
\includegraphics[trim = 10mm 5mm 20mm 5mm,clip=true, width=3.1cm,height = 1.6cm]{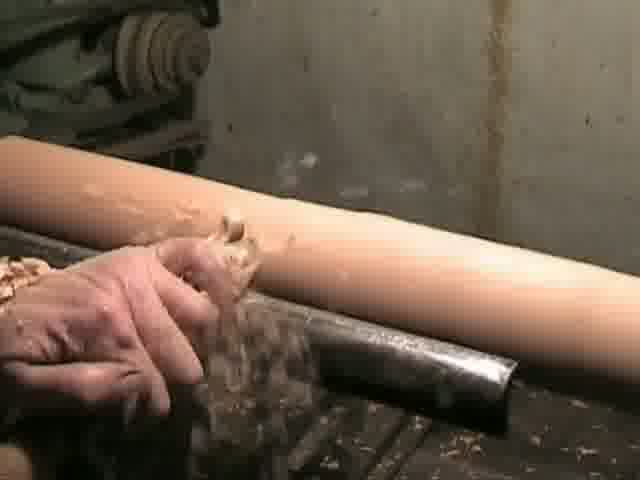}\\E001&E002&E003&E004&E005
\end{tabular}
\caption{Sample images from the five events in the MED11 dataset: (E001) attempting a board trick; (E002) feeding an animal; (E003) landing a fish; (E004) wedding ceremony; (E005) working on a woodworking project.}
\label{fig:F5}
\end{figure*}

The TRECVID MED11 was used for evaluating our performance on event recognition. We use the first five events containing 813 video clips with millions of frames, representing complex activities and events, see Fig. (\ref{fig:F5}) for sample images.  We extract local features using HOG3D \cite{Klaser08}. To represent videos using local features, we apply a common bag-of-words model with a 1000-word codebook. Each frame is represented as a histogram of occurrences of the codebook elements. Because this dataset requires large memory, we run it on 24 cores machine (Intel(R) Xeon(R) CPU  X5650  @2.67GHz), with memory 48GB. 

\begin{figure}[h!]
\begin{center}
\begin{tabular}{cc}
\includegraphics[trim = 40mm 84mm 40mm 85mm,clip=true, width=6.0cm]{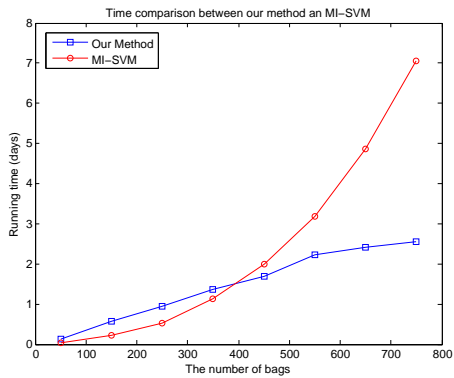}
\end{tabular}
\end{center}
\caption{The running time comparison. We randomly sample \#bags from the 813 total bags and test the running time of two methods respectively.}
\label{fig:Fig9}
\end{figure}

\begin{table*}[ht!]
\begin{center}
\begin{tabular}{lcccccccc}
\hline
\multicolumn{1}{l}{\multirow{2}{*}{Method}}& \multicolumn{7}{c}{Accuracy (\%) comparison on MED11 dataset} \\
\cline{2-8}
& E001 & E002 & E003 & E004 & E005  & Avg. Accuracy & Time (Days)\\
\cline{1-8}
MI-SVM \cite{Andrews02}& $37.20$ & $43.94$ & 28.80 & 51.64 & 67.34 & 45.78 & $\ge7$ days \\
GMIL (this paper) & 57.20 & 47.70 & 71.90 & 86.70 & $76.70$ & \textbf{68.04} & $1\sim2$ days \\
\hline
\end{tabular}
\caption{Accuracy comparison on five events for different methods (10 fold cross validation) on 24 cores machine (Intel(R) Xeon(R) CPU 702 X5650 @2.67GHz), with 48GB RAM. It demonstrates that our method can yield competitive results. }
\label{Tab3}
\end{center}
\end{table*}
In this experiment, we only have video level labels. Thus, we treat event recognition as a MIL problem. Each video can be seen as a bag, and its frames in the corresponding bag as instances. To evaluate the performance of our method, we use MI-SVM as baseline because structural SVM (svmlight \footnote{\url{http://svmlight.joachims.org/}} or liblinear SVM) requires linear time for training (just linear kernel). Thus it is faster compared to other methods; for example, mi-Graph has the time complexity $O(N^2(m+n)^2)$. In the experiment, MI-SVM deploys liblinear\footnote{\url{http://www.csie.ntu.edu.tw/~cjlin/liblinear/}} for training.
For evaluation, we take one-vs-all strategy. 
For our method, we set $K=50$, $\textbf{p} = 3$, and $\textbf{q}=9$ to train on the unbalanced dataset. The results in Table (\ref{Tab3}) demonstrate that our method yields competitive results. We also compare the running time between our method and MI-SVM, see Fig. \ref{fig:Fig9} for details.





\section{Conclusion}
In this paper, we propose a greedy multiple instance learning method that leverages codebook learning and nearest neighbor voting. Instead of maximizing a likelihood function, we take a greedy strategy to maximize the histogram ratio between positive bags and negative bags. By learning codebooks for both targets and negative centers, we can use them to classify novel bags based on nearest neighbor strategy. The primary contribution of this paper is to maximize the density ratio to speed up the learning process. Another contribution is learning both targets and negative candidate centers to reduce false positives. Experimental results show that our method is significantly faster and effective compared to the state of the art. In future work, we will consider weight to cluster instances to learn the codebooks. For example, we can use Mahalanobis distance as a distance measure in K-means. We also plan to investigate spectral clustering or kernel k-means methods to learn the codebooks. It is worth mentioning that we can use a more theoretical and more sophisticated kernel function for kernel density estimation in order to improve classification accuracy.

{\small
\bibliography{egbib}\setlength{\itemsep}{-2mm}
\bibliographystyle{splncs}
}

\section*{Appendices}\label{app}
\begin{proof}
 The Diverse Density model (a noise-or model) maximizes the following equation:
\begin{align}
&\argmax_{x} {Pr(x=t \mid B_{1}^{+},...,B_{n}^{+})}{Pr(x=t \mid B_{1}^{-},...,B_{m}^{-})} &\nonumber\\
=&\argmax_{x}  {\prod_{i=1}^n Pr(x=t \mid B_{i}^{+})}{\prod_{i=1}^m Pr(x=t \mid B_{i}^{-})} & \nonumber\\  
= &\argmax_{x} \underset{\text{${B^{+}}$ likelihood term}}{\underbrace {\prod_{i=1}^n \Big(1-\prod_{j} \big(1-Pr(x=t \mid B_{ij}^{+} \big)\Big)}} &\nonumber\\ 
& \quad \quad\quad \times \underset{\text{${B^{-}}$ likelihood term}}{\underbrace {\prod_{i=1}^m \prod_{j} \big (1- Pr(x=t \mid B_{ij}^{-})\big)}} &
\label{eq:Eq0}
\end{align}
Note that the left likelihood term and the right term share a common term $\big(1- Pr(x=t \mid B_{ij}^{*})\big)$ (the symbol * can be + or -), except one instance is from positive training bags, while the other instance is from negative training bags. Thus, to maximize Eq. (\ref{eq:Eq0}) means that we need to minimize $\prod_{j}\big(1- Pr(x=t \mid B_{ij}^{+})\big)$ and maximize $\prod_{j}\big(1- Pr(x=t \mid B_{ij}^{-})\big)$ simultaneously. Equivalently, we can further minimize the following ratio:
\begin{eqnarray}\label{eq:Eq0_1}
\begin{aligned}
 &  \underset{x}{\operatorname{arg\,min}}  \frac {\prod_{i=1}^n \prod_{j} \big(1-Pr(x=t \mid B_{ij}^{+} )\big)} {\prod_{i=1}^m \prod_{j} \big (1- Pr(x=t \mid B_{ij}^{-})\big)} &\\
 =\quad &  \underset{x}{\operatorname{arg\,max}}  \frac {\prod_{i=1}^n \prod_{j} Pr(x=t \mid B_{ij}^{+} )} {\prod_{i=1}^m \prod_{j}  Pr(x=t \mid B_{ij}^{-})} &\\
\ge \quad &  \underset{x}{\operatorname{arg\,max}}  \frac {\sum_{i=1}^n \sum_{j} Pr(x=t \mid B_{ij}^{+} )} {\sum_{i=1}^m \sum_{j}  Pr(x=t \mid B_{ij}^{-})} &\\
\end{aligned}
\end{eqnarray}
The last two equations can be derived from theorem \ref{th1}. Thus, the density ratio Eq. (\ref{eq:Eq2}) is the lower bound of the Diverse Density model. Intuitively, if one of the instances in a positive bag is close to $x$, then $Pr(x=t \mid B_{ij}^{+})$ is highly peaked, and Eq. (\ref{eq:Eq1}) will have high value, which guarantees high Diverse Density in Eq. (\ref{eq:Eq0}). Further, if every positive bag has an instance close to $x$ and no negative bags are close to $x$, then $x$ will have high value in Eq. (\ref{eq:Eq1}), as well as high Diverse Density in Eq. (\ref{eq:Eq0}). 
In other words, to maximize the density ratio can directly maximize the DD model.
\end{proof}

\end{document}